\pdfoutput=1

\documentclass[11pt]{article}

\usepackage{emnlp2021}

\usepackage{times}
\usepackage{latexsym}

\usepackage{enumitem}
\newlist{researchquestions}{enumerate}{10}
\setlist[researchquestions]{label*=\textbf{RQ\arabic*}}

\usepackage[T1]{fontenc}

\usepackage[utf8]{inputenc}

\usepackage{microtype}

\usepackage{xurl}
\usepackage{multirow}
\usepackage{subcaption}
\usepackage{array}

\usepackage{listings}
\usepackage{enumitem}
\usepackage{cleveref}
\usepackage{soul}
\usepackage{adjustbox}
\usepackage{float}
\usepackage{appendix}
\usepackage{booktabs}
\usepackage{tabularx}

\makeatletter
\def\mathcolor#1#{\@mathcolor{#1}}
\def\@mathcolor#1#2#3{%
  \protect\leavevmode
  \begingroup
    \color#1{#2}#3%
  \endgroup
}
\definecolor{green}{rgb}{0.05, 0.9, 0.05}

\definecolor{redbw}{HTML}{d7191c}
\definecolor{greenbw}{HTML}{1c8036}
\definecolor{bluebw}{HTML}{2b83ba}

\newcommand\mup[1]{\footnotesize{\color{greenbw}#1\color{black}}}
\newcommand\mdo[1]{\footnotesize{\color{redbw}#1\color{black}}}
\newcommand\msa[1]{\footnotesize{#1}}

\usepackage{amssymb}

\title{Open Aspect Target Sentiment Classification \\ with Natural Language Prompts}

\author{Ronald Seoh\Thanks{ Equal contribution.}\hspace{0.15cm}\textsuperscript{1}\hspace{0.25cm}Ian Birle\footnotemark[1]\hspace{0.15cm}\textsuperscript{1}\hspace{0.25cm}Mrinal Tak\footnotemark[1]\hspace{0.15cm}\textsuperscript{1}\hspace{0.25cm}Haw-Shiuan Chang\footnotemark[1]\hspace{0.15cm}\textsuperscript{1} \\ \textbf{Brian Pinette\textsuperscript{2}\hspace{0.25cm}Alfred Hough\textsuperscript{2}} \\ \textsuperscript{1 }University of Massachusetts Amherst \hspace{0.25cm}\textsuperscript{2 }Lexalytics, Inc.  \\
  \texttt{\{bseoh, ibirle, mtak, hschang\}@cs.umass.edu} \\
  \texttt{\{brian.pinette, al.hough\}@lexalytics.com}}

\begin{document}

\maketitle

\begin{abstract}
For many business applications, we often seek to analyze sentiments associated with any arbitrary aspects of commercial products, despite having a very limited amount of labels or even without any labels at all. However, existing aspect target sentiment classification (ATSC) models are not trainable if annotated datasets are not available. Even with labeled data, they fall short of reaching satisfactory performance. To address this, we propose simple approaches that better solve ATSC with natural language prompts, enabling the task under zero-shot cases and enhancing supervised settings, especially for few-shot cases. Under the few-shot setting for SemEval 2014 Task 4 laptop domain, our method of reformulating ATSC as an NLI task outperforms supervised SOTA approaches by up to 24.13 accuracy points and 33.14 macro F1 points. Moreover, we demonstrate that our prompts could handle \emph{implicitly} stated aspects as well: our models reach about 77\% accuracy on detecting sentiments for aspect categories (e.g., food), which do not necessarily appear within the text, even though we trained the models only with explicitly mentioned aspect terms (e.g., fajitas) from just 16 reviews — while the accuracy of the no-prompt baseline is only around 65\%.
\end{abstract}

\section{Introduction}

\begin{figure}[t]
\centering
\includegraphics[width=\columnwidth, clip, trim=1cm 3cm 1.1cm 1cm]{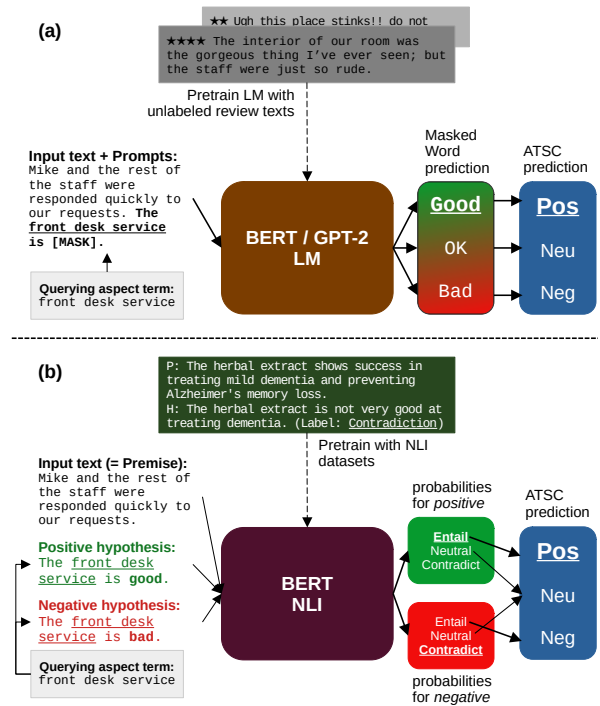}
\caption{An overview of our prompt-based models for ATSC. (a) BERT / GPT-2 LM is pretrained on unlabeled review texts and we convert ATSC into a language modeling task during training and testing. (b) BERT NLI is pretrained on NLI datasets and we convert ATSC into two entailment tasks.}
\label{fig:overview}
\end{figure}

Measuring targeted sentiments from text toward certain aspects or subtopics has immediate commercial value. For example, a hotel chain might want to base their business decisions on the proportion of customer reviews being positive toward their room cleanliness and front desk services. Manually reading through thousands of reviews would be prohibitively expensive, calling for automated solutions.

Adopting existing supervised models for aspect target sentiment classification (ATSC) \citep{pontiki-etal-2014-semeval} appears to be an obvious choice at first glance. However, the accuracy of these models in practice are often unsatisfactory due to the lack of labeled data from domains of interest. We could direct practitioners to annotate few domain-specific review texts, but ATSC models would need to generalize from those limited examples and return accurate predictions upon seeing any arbitrary aspects and sentiments. Even if we could get hold of more data, it would be difficult to collect enough labels to handle all aspect-sentiment cases that customers could ever query.

The way people would write customer reviews inspires our solutions to this label scarcity issue. For example, consider two plausible sentences from hotel reviews: \emph{Mike and the rest of the staff were very polite and responded quickly to our requests. The front desk service is excellent.} Supposing we only observe the first one, if an ATSC model could tell that the first \emph{entails} the second, or the second \emph{naturally follows} the first, we could determine that the customer feels \ul{positive} about the front desk service. 

Based on this intuition, we propose two ATSC models based on natural language prompts: our first method appends cloze question prompts (e.g., \textit{The service is }) to the review text and predicts how likely it is to observe \textit{good}, \textit{bad}, and \textit{ok} as the next word. The second method treats the review as a premise and the prompt sentence with the sentimental next word as a hypothesis, and predicts whether the review entails the prompt. With pretraining, these prompts give us natural ways of leveraging more abundant unlabeled reviews or large natural language inference (NLI) datasets to overcome the scarcity of labeled data. An overview of our methods is provided in \autoref{fig:overview}.

Most of previous ATSC efforts focus on fully supervised training, assuming enough in-domain labels on relevant aspects. Going beyond this typical setting, we also test our model on more ``open’’ situations where reviews and aspects for testing are less similar to the training examples.

Experimental results show that our methods are not only capable of producing reasonable predictions without any labeled data (zero-shot) but also constantly outperform SOTA no-prompt baselines under supervised settings. Moreover, our prompt-based models are robust to \emph{domain shifts}: after training with aspect-based sentiments in one domain, we observe that the models can accurately predict sentiments associated with aspect terms appearing in reviews from a completely unseen domain. Also, our models that are trained on explicitly mentioned aspect terms could generalize well to implied \emph{categorical} aspects.\footnote{All the program codes used to produce results presented in this paper are available at \url{https://link.iamblogger.net/atscprompts}.}

\section{Related Work}

\subsection{Aspect-based Sentiment Analysis}

\citet{hu2004mining} is one of the first academic work to discuss the task of analyzing opinions targeted towards different aspects or topics within the text. \citet{pontiki-etal-2014-semeval} starts the current line of research on aspect-based sentiment analysis (ABSA), with their benchmark datasets of customer reviews for restaurants and laptops.

To overcome small training dataset sizes, recent developments for ABSA involve some combination of unlabeled domain text pretraining and intermediate task finetuning. \citet{xu-etal-2019-bert} and \citet{chi19} use multi-task loss functions to finetune BERT using SQuAD question answering datasets that contain a range of domains and task-related knowledge, to offset the small size of SemEval 2014 ABSA datasets. \citet{rietzler19} provides detailed analysis on \emph{cross-domain adaptation}, where they find that end task finetuning with the domain different from the evaluation domain still achieves performance comparable to the SOTA results using in-domain labels.

\subsection{Natural Language Prompts}

There has been a number of recent papers on using prompts — additional sentences appended to the original input text — to direct language models to perform different tasks, exploiting the knowledge they have acquired during their original pretraining. One of the earliest examples of such efforts is \citet{radford2019language}, where they measured their GPT-2 model's performance on downstream tasks by feeding in task descriptions as prompts, without any finetuning at all. Since then, a number of previous work has leveraged prompts for the tasks such as question answering \citep{lewis2019unsupervised} and commonsense reasoning \citep{shwartz2020unsupervised}. We also note that some previous work on prompt-based learning methods have included sentence-level sentiment classification, which measures sentiment from the entire input text, as part of their evaluation \citep{shin2020autoprompt, gao-etal-2021-making}.

Many recent works, including ours, follow the format of \emph{cloze questions} to design prompts as first suggested by \citet{schick2020exploiting}. We design prompts to include masked tokens that need to be filled in, and the predictions for the masks serve as the outputs for the original task. 

\section{Methods}

To maximally leverage the large data resources of unlabeled review texts and NLI datasets, we propose two ways of reformulating ATSC: the first converts ATSC into next/masked word prediction; the second transforms the task into NLI entailment predictions.

\subsection{ATSC as Language Modeling}

In order to elicit abilities to perform ATSC from language models (LMs), it was essential that prompt sentences should be similar to what one would typically write to express their sentiment. Hence, we came up with the following set\footnote{In our experiments, all three prompts perform similarly, especially in few-shot cases; this suggests that the performance of our models are not overly sensitive to the wording of properly chosen prompts. We present the performance comparison between our prompts in \autoref{appendix:prompts_comparison}.} of cloze question prompts \citep{schick2020exploiting} that are aspect dependent, and appended them to input texts:

\small
\begin{itemize}
\setlength\itemsep{0.01mm}
\item \texttt{I felt the \{aspect\} was [MASK].} 
\item \texttt{The \{aspect\} made me feel [MASK].}
\item \texttt{The \{aspect\} is [MASK].}
\end{itemize}
\normalsize

\noindent where \texttt{\{aspect\}} is the placeholder for the querying aspect term, and \texttt{[MASK]} represents the masked word for BERT \citep{devlin2019bert} and the next word for GPT-2 \citep{radford2019language}. Then, we let the probability of predicting \textsf{positive}, \textsf{neutral}, and \textsf{negative} sentiment be proportional to the probability of predicting \texttt{good}, \texttt{ok}, and \texttt{bad}, respectively.

Publicly available pretrained weights for BERT and GPT-2 have already been trained on large general corpora, but they might not include enough sentences from our testing domains, such as laptop reviews. To produce more accurate predictions on \texttt{[MASK]}, we further pretrain LMs with the in-domain review texts. For BERT, we modify the original random masking scheme to mask only adjectives, nouns, and proper nouns because the words are more likely to indicate the sentiments of the sentence. For GPT-2, we use the original causal LM (CLM) objective. To measure the effectiveness of the prompts, our baselines without prompts also receive identical pretraining.\footnote{We compare the results between original weights and our further pretrained ones in \autoref{appendix:vanilla_bert}.}

When the labeled data for ATSC is available, we convert the training labels to \texttt{good}, \texttt{ok}, and \texttt{bad} and finetune all the parameters of LMs, including the encoders and embeddings of words in the prompts. During the training and testing, other candidates for \texttt{[MASK]} are ignored.

\subsection{ATSC as NLI}

We first set the input review text as a premise. We predict the scores for \textsf{positive}, \textsf{negative}, and \textsf{neutral} sentiment using the entailment probabilities from a NLI model as follows: we create positive and negative hypotheses by populating prompts with corresponding label words (e.g., \small\texttt{``The \{aspect\} is good; The \{aspect\} is bad.''}\normalsize). We get the scores for \textsf{positive} and \textsf{negative} sentiment by obtaining entailment probabilities with each of the hypotheses; For the \textsf{neutral} class, we average neutral probabilities (from NLI) for the two hypotheses. Our method enables zero-shot ATSC, which previously had not considered by previous efforts that leveraged NLI for sentiment analysis and text classification tasks \citep{yin2019benchmarking,chi19,wang2021entailment}.

We use the BERT-base model pretrained on the MNLI dataset \citep{N18-1101}, and none of the unlabeled review texts are utilized for pretraining. We apply a softmax layer on top of the logits to normalize the prediction scores of the three classes into probabilities, in order to finetune models with cross-entropy loss when labeled data is available.  

\begin{table*}[t]
    \centering
    \begin{subtable}[h!]{0.88\textwidth}
    \begin{adjustbox}{width=\textwidth}
    \begin{tabular}{ccc|cc|cc|cc|cc|cc}
        \multirow{3}{*}{\textbf{Model}} & \multicolumn{12}{c}{\textbf{Number of Training Examples}} \\
                                           & \multicolumn{2}{c}{Zero} & \multicolumn{2}{c}{16} & \multicolumn{2}{c}{64} & \multicolumn{2}{c}{256} & \multicolumn{2}{c}{1024} & \multicolumn{2}{c}{Full (1850)} \\
                                                          & Acc   & MF1   & Acc   & MF1   & Acc   & MF1   & Acc   & MF1   & Acc   & MF1   & Acc   & MF1 \\
        \midrule
        {\centering BERT-ADA}                             & --    & --    & --    & --    & --     & --   & --    & --    & --    &   --  & 79.19\(\dagger\)  & 74.18\(\dagger\) \\
        \midrule
        {\centering BERT [CLS]}                           & --    & --    & \ul{48.75} & \ul{34.92} & 60.63      & \ul{49.43} & 72.35      & 64.31      & \ul{76.87}   & \ul{71.22} & 80.06 & 75.08\\
        {\centering BERT NSP}                          & --    & --    & 48.24      & 31.35      & \ul{60.91} & 49.27      & \ul{72.38} & \ul{64.64} & 76.77        &   71.12            & \ul{80.25} & \ul{75.46}\\
        \specialrule{.2em}{.3em}{.3em}
        \multirow{2}{*}{\centering BERT LM}            & \textbf{63.58} & 46.17 & 69.05 & 58.60 & 72.80 & 65.54 & \textbf{76.59} & 70.65 & \textbf{79.30} & 74.80 & \textbf{81.10} & 76.83 \\
                                                          & \msa{--} & \msa{--} & \mup{+20.30}* & \mup{+23.68}* & \mup{+11.89}* & \mup{+16.11}* & \mup{\textbf{+4.21}}* & \mup{+6.01}* & \mup{\textbf{+2.43}}* & \mup{+3.58}* & \mup{\textbf{+0.85}}* & \mup{+1.37}*\\
        \midrule
        \multirow{2}{*}{\centering GPT-2 LM}           & 60.45 & 39.59 & 68.94 & 56.71 & 71.54 & 63.69 & 76.48 & 70.89 & 79.02 & \textbf{74.88} & 80.73 & \textbf{77.13}\\
                                                          & \msa{--} & \msa{--} & \mup{+20.19}* & \mup{+21.79}* & \mup{+10.63}* & \mup{+14.26}* & \mup{+4.10}* & \mup{+6.25}* & \mup{+2.15}* & \mup{+3.66}* & \mup{+0.48} & \textbf{\mup{+1.67}}*\\
        \specialrule{.2em}{.3em}{.3em}
        \multirow{2}{*}{\centering BERT NLI}             & 58.93 & \textbf{54.91} & \textbf{72.88} & \textbf{68.06} & \textbf{74.95} & \textbf{70.84} & 76.22 & \textbf{71.65} & 77.42 & 73.52 & 77.58 & 73.18\\
                                                          & \msa{--} & \msa{--} & \mup{\textbf{+24.13}}* & \mup{\textbf{+33.14}}* & \mup{\textbf{+14.04}}* & \mup{\textbf{+21.41}}* & \mup{+3.84}* & \mup{\textbf{+7.01}}* & \mup{+0.55} & \mup{+2.30} & \mdo{-2.67} & \mdo{-2.28}\\
        \bottomrule
    \end{tabular}
    \end{adjustbox}
    \caption{Laptops}
    \end{subtable}
    
    \hspace{40em}
    
    \begin{subtable}[h!]{0.88\textwidth}
    \begin{adjustbox}{width=\textwidth}
    \begin{tabular}{ccc|cc|cc|cc|cc|cc}
                                           & \multicolumn{2}{c}{Zero} & \multicolumn{2}{c}{16} & \multicolumn{2}{c}{64} & \multicolumn{2}{c}{256} & \multicolumn{2}{c}{1024} & \multicolumn{2}{c}{Full (3602)} \\
                                                          & Acc   & MF1   & Acc   & MF1   & Acc   & MF1   & Acc   & MF1   & Acc   & MF1   & Acc   & MF1 \\
        \midrule
        {\centering BERT-ADA}                             & --    & --    & --    & --    & --     & --   & --    & --    & --    &   --  & 87.14\(\dagger\) & 80.05\(\dagger\) \\
        \midrule
        {\centering BERT [CLS]}                           & --    & --    & 59.89      & \ul{34.50} & 73.00      & 50.79      & \ul{79.45} & 64.70      & 83.48      & 73.62 & 86.77 & 79.33\\
        {\centering BERT NSP}                          & --    & --    & \ul{61.05} & 32.46      & \ul{74.73} & \ul{53.00} & 79.34      & \ul{65.51} & \ul{83.61} & \ul{74.15} & \ul{87.09} & \ul{79.98}\\
        \specialrule{.2em}{.3em}{.3em}
        \multirow{2}{*}{\centering BERT LM}            & 70.86 & 48.17           & 71.99        & 56.65       & 77.79        & 63.30        & 81.10       & 69.27       & 85.12 & 76.60 & \textbf{87.50} & \textbf{80.78}\\
                                                          & \msa{--} & \msa{--} & \mup{+10.94}* & \mup{22.15}* & \mup{+3.06}*  & \mup{+10.30}* & \mup{+1.65}* & \mup{+3.76}* & \mup{+1.51}* & \mup{+2.45}* & \mup{\textbf{+0.41}} & \mup{\textbf{+0.80}}\\
        \midrule
        \multirow{2}{*}{\centering GPT-2 LM}           & \textbf{71.40} & 45.53          & \textbf{75.41}        & 60.06        & 79.30       & 65.49       & \textbf{82.27}       & 71.62       & \textbf{85.28} & \textbf{77.38} & 86.99 & 80.02\\
                                                          & \msa{--} & \msa{--} & \textbf{\mup{+14.36}}* & \mup{+25.56}* & \mup{+4.57}* & \mup{+12.49}* & \mup{\textbf{+2.82}}* & \mup{+6.11}* & \mup{\textbf{+1.67}}* & \mup{\textbf{+3.23}}* & \mdo{-0.1} & \mup{+0.04}\\
        \specialrule{.2em}{.3em}{.3em}
        \multirow{2}{*}{\centering BERT NLI}             & 61.79 & \textbf{57.93}           & 74.74       & \textbf{65.58}        & \textbf{79.33}        & \textbf{69.44}        & 81.24       & \textbf{71.94}       & 83.07 & 74.52 & 85.07 & 77.53\\
                                                         & \msa{--} & \msa{--} & \mup{+13.69}* & \mup{\textbf{+31.08}}* & \mup{\textbf{+4.60}}*& \mup{\textbf{+16.44}}* & \mup{+1.79}* & \mup{\textbf{+6.43}}* & \mdo{-0.54} & \mup{+0.37} & \mdo{-2.02} & \mdo{-2.45}\\
        \bottomrule
    \end{tabular}
    \end{adjustbox}
    \caption{Restaurants}
    \end{subtable}
    \caption{Results of our methods and baselines. Acc and MF1 refer to accuracy and macro F1, respectively. We use five random seeds for each of the prompts and baselines, and average their scores. We averaged the performance of our models across all three prompts. Please see \autoref{appendix:prompts_comparison} for performance comparison between the prompts. Boldfaces indicate the best performance given the same number of labels, and the best baseline scores are underlined. \(\dagger\) BERT-ADA results are taken directly from \citet{rietzler19}. * indicates an increase over the baseline with significance level .05 using a two mean z-test.}
    \label{tab:results}
\end{table*}

\section{Experiments}

In the experiments, we test the generalization capability of our methods on more real world-like conditions where there are far fewer training examples similar to the testing examples. Using ATSC datasets from SemEval 2014 Task 4 Subtask 2 \citep{pontiki-etal-2014-semeval}, we evaluate our models on the full spectrum of in-domain training data sizes covering the zero-shot and full-shot (i.e., fully supervised) cases. Similar to the settings of \citet{scao2021many}, we train our models with randomly re-sampled training sets of sizes \{Zero, 16, 64, 256, 1024, Full\}.

Furthermore, we conduct \emph{cross-domain} evaluation where we train the models on restaurant reviews and test on laptop reviews, and vice versa. Finally, we train the models on ATSC and test them on aspect \emph{category} sentiment classification (ACSC), another ABSA variant, to evaluate the robustness to an unseen querying aspect distribution. Unlike aspect terms, these categories such as \emph{food, service, price, and ambience} usually do not explicitly appear within the text, but \emph{implicitly} stated through aspect terms or overall context.\footnote{Please refer to \autoref{appendix:dataset_info} for ATSC/ACSC dataset statistics and preprocessing.}

\subsection{Baselines}

We compared our prompt-based methods with two common strategies of utilizing BERT for classification tasks: 1) the last hidden state of \texttt{[CLS]} token (BERT [CLS]), and 2) the NSP head of BERT (BERT NSP). We note that the architecture of BERT NSP is equivalent to BERT-ADA \citep{rietzler19}, currently the top-performing BERT-based model for ATSC which we show their reported full-shot performance for reference.

\subsection{Results}

\textbf{Prompts constantly outperform the no-prompt baselines.} We can see in \autoref{tab:results} that for both target domains, our prompt-based BERT models outperform the no-prompt baselines in all cases. Especially for few-shots, we achieve larger performance gains as fewer labels are available. We note that BERT NLI does particularly well in 16 to 256 shots for laptops, with noticeably higher accuracy and macro F1 (MF1) than other prompt models.\footnote{With the help of more abundant neutral examples in MNLI, \autoref{appendix:error_analysis} suggests that BERT NLI is particularly better at detecting neutral sentiment, subsequently leading to better MF1 scores.}

We emphasize again that our NLI models are only trained on the MNLI dataset, which makes them particularly preferable when in-domain text (shopping reviews) is not readily available.

Lastly, we observe that our methods achieve good performances in the zero-shot cases, significantly outperforming the baselines that are trained on 16 samples, further showing its practicality.

\begin{table}[tb]
    \centering
    \begin{adjustbox}{width=\columnwidth}
    \begin{tabular}{cc|cc|cc}
        \toprule
        \multirow{2}{*}{\textbf{Model}}        & \multirow{2}{*}{\textbf{In/Cross}}  & \multicolumn{2}{c}{\textbf{16}}    & \multicolumn{2}{c}{\textbf{Full}} \\    
                                             &                    &  \msa{Restaurants} & \msa{Laptops} & \msa{Restaurants} & \msa{Laptops} \\
        \midrule
        \multirow{2}{*}{\centering BERT NSP} & In                 & 61.05       & 48.24                & 87.09       & 80.25 \\
                                             & Cross              & 49.55       & 47.46                & 81.29       & 78.56 \\
        \specialrule{.2em}{.3em}{.3em}
        \multirow{2}{*}{\centering BERT LM}  & In                 & 71.99       & 69.05                & 87.50       & 81.10 \\
                                             & Cross              & 75.21       & 68.17                & 81.27       & 79.03 \\
        \midrule
        \multirow{2}{*}{\centering BERT NLI} & In                 & 74.74       & 72.88                & 85.07       & 77.58\\
                                             & Cross              & 77.45       & 70.43                & 80.35       & 76.61\\
        \bottomrule
    \end{tabular}
    \end{adjustbox}
    \caption{Accuracies of BERT NSP, LM, and NLI trained with in-domain and cross-domain data.}
    \label{tab:nli-domains}
\end{table}

\textbf{Prompts can utilize cross-domain data more effectively.} As shown in \autoref{tab:nli-domains}, the prompt models with 16-shot cross-domain training achieve better performance than both in- and out-domain BERT NSP. It is also interesting to note that cross-domain have even exceeded in-domain for the restaurants domain. This suggests that our methods might have the potential to be particularly more adaptable to arbitrary domains under low-resource settings, which we plan to explore further in future research.

\begin{table}[tb]
    \centering
    \begin{adjustbox}{width=0.70\columnwidth}
    \begin{tabular}{ccccc}
        \toprule
        \multirow{2}{*}{\textbf{Model}}        & \multicolumn{2}{c}{\textbf{16}}          & \multicolumn{2}{c}{\textbf{Full}} \\    
                                               &  Acc & MF1 & Acc & MF1  \\
        \midrule
        BERT NSP                               &  64.73  &  33.22 & 82.45 & 70.91\\
        \midrule
        BERT LM                                &  76.67  &  56.77 & 84.31 & 74.14\\
        BERT NLI                               &  66.92  &  58.18 & 67.42 & 59.24\\
        \bottomrule
    \end{tabular}
    \end{adjustbox}
    \caption{Performance on ACSC without any extra training. Refer to \autoref{appendix:more_on_acsc} for the results with other training set sizes.}
    \label{tab:subtask4}
\end{table}

\textbf{Prompts can better recognize \emph{implicit} aspects.} We can see from \autoref{tab:subtask4} that the BERT LM model trained with merely 16 examples achieves about 77\% accuracy on ACSC, while it was never trained in terms of aspect categories at all. BERT NSP, the no-prompt baseline, achieves around 65\%. This result suggests that our prompt-based models have also acquired some abilities to recognize aspects that are implied or worded differently from the querying aspect term. Such abilities could also make our prompt models more desirable for potential real-life applications. We note that BERT NLI performs rather poorly, particularly under full-shot. As it hadn't seen in-domain texts during pretraining, we suspect that it cannot fully recognize related domain-specific words.

\section{Conclusion and Future Work}

In this paper, we examined our prompt-based ATSC models leveraging LM and NLI under zero-shot, few-shot, and full supervised settings. We observe a significant amount of improvements over the no-prompt baselines in nearly all configurations we have tested. In particular, we find that our NLI model performs well with lower amounts of training data, while the BERT LM model does better when more labels are available. In addition, we have seen that it could effectively utilize cross-domain labels and recognize implicit aspects, suggesting that it would potentially be more applicable in real-life scenarios.

For future work, one direction is to adapt our aspect-dependent prompts to the models that jointly perform aspect term extraction and sentiment classification, such as \citet{luo-etal-2020-grace}. Secondly, we could explore potential ways of combining ATSC, masked/next word prediction, and NLI into a unified task in order to take the full advantage of both our unlabeled text and NLI pretraining. Lastly, it would be an interesting analysis to determine whether there are any strong linguistic patterns among correct or incorrect predictions that each of our models make — such findings could allow us to have more detailed insights into the potential behaviors of our prompt-based models.

\section*{Acknowledgments}

We thank our anonymous reviewers for providing valuable feedback. This work was supported by the Center for Data Science at the University of Massachusetts Amherst, under the Center's industry mentorship program. We would like to express our gratitude to Professor Mohit Iyyer and Professor Andrew McCallum for suggesting constructive discussions throughout the course of our project. We also would like to thank Paul Barba of Lexalytics for setting the project's initial research directions. Last but not least, we appreciate all the efforts Xiang Lorraine Li and Rico Angell have put in to manage the program participants and provide various practical assistance.

This work was produced in part using high performance computing equipment obtained under a grant from the Collaborative R\&D Fund, managed by the Massachusetts Technology Collaborative.

\bibliography{anthology,custom_new}
\bibliographystyle{acl_natbib}

\clearpage
\newpage
\appendix

\section{Dataset Information}
\label{appendix:dataset_info}

\subsection{SemEval 2014 Task 4 dataset} 

\paragraph{Aspect Target Sentiment Classification (ATSC, Subtask 2)} The dataset released by \citet{pontiki-etal-2014-semeval} is one of the most popular benchmark datasets for ATSC used in the literature, which contains English review sentences from two target domains, laptops, and restaurants. We measure the performance of our models and baselines on the test splits of each domain.

Labels for the sentiments are limited to \textsf{positive}, \textsf{negative}, \textsf{neutral}, and \textsf{conflict}. \textsf{neutral} refers to the case where the opinion towards the aspect is neither positive nor negative, and \textsf{conflict} is for both positive and negative sentiments being expressed for \emph{one} aspect.

To make our work comparable with previous efforts \citep{xu-etal-2019-bert, rietzler19}, we use the following preprocessing procedure:

\begin{enumerate}
    \item Reviews with \textsf{conflict} labels were removed. They have been usually ignored due to having a very small number of examples.
    \item Multiple aspect-sentiment labels within one text piece were split up into different data points.
\end{enumerate}

Dataset statistics after preprocessing are provided in \autoref{tab:semeval_stat}.

\begin{table}[H]
    \centering
    \begin{adjustbox}{width=0.8\columnwidth}
    \begin{tabular}{ccccccc}
        \toprule
        \multirow{2}{*}{Class} & \multicolumn{2}{c}{\textbf{Restaurant}} & \multicolumn{2}{c}{\textbf{Laptop}}\\
                               & Train & Test                            & Train & Test\\
        \midrule
        Positive               & 2164 & 728                              & 987 & 341\\
        Negative               & 645  & 196                              & 866 & 128 \\
        Neutral                & 496  & 196                              & 460 & 169 \\
        \midrule
        All                    & 3602 & 1120                             & 1850 & 638 \\
        \bottomrule
    \end{tabular}
    \end{adjustbox}
    \caption{SemEval 2014 dataset statistics after preprocessing.}
    \label{tab:semeval_stat}
\end{table}

\paragraph{Aspect Category Sentiment Classification (ACSC, Subtask 4)} For this task, the labeled data is available only for the restaurant domain. While the class labels are the same as ATSC, this task has predefined aspect categories: \textsf{food, price, service, ambience, anecdotes/miscellaneous}. We only use the test split with 973 examples, containing 657 positives, 222 negatives, and 94 neutrals. The train split for ACSC is never used in any manner throughout our experiments.

\subsection{LM Pretraining Corpora}

We note the following sources of unlabeled review texts to further pretrain BERT and GPT-2 language models for two target domains of SemEval 2014 Task 4:

\begin{enumerate}
    \item \textbf{Amazon Review Data} \citep{ni-etal-2019-justifying} is the collection of customer reviews extracted from the online shopping website Amazon. We used 20,994,353 reviews written for the products from the electronics category. The LMs pretrained with this collection are used to target the ATSC \ul{laptop} domain.
    \item \textbf{Yelp Open Dataset}\footnote{\url{https://www.yelp.com/dataset}} consists of over 8 million business reviews. We extracted the reviews associated with restaurants (2,152,007 reviews). The LMs pretrained with this collection are used to target the ATSC \ul{restaurants} domain.
\end{enumerate}

\begin{figure*}[t!]
\centering
\includegraphics[width=0.9\linewidth]{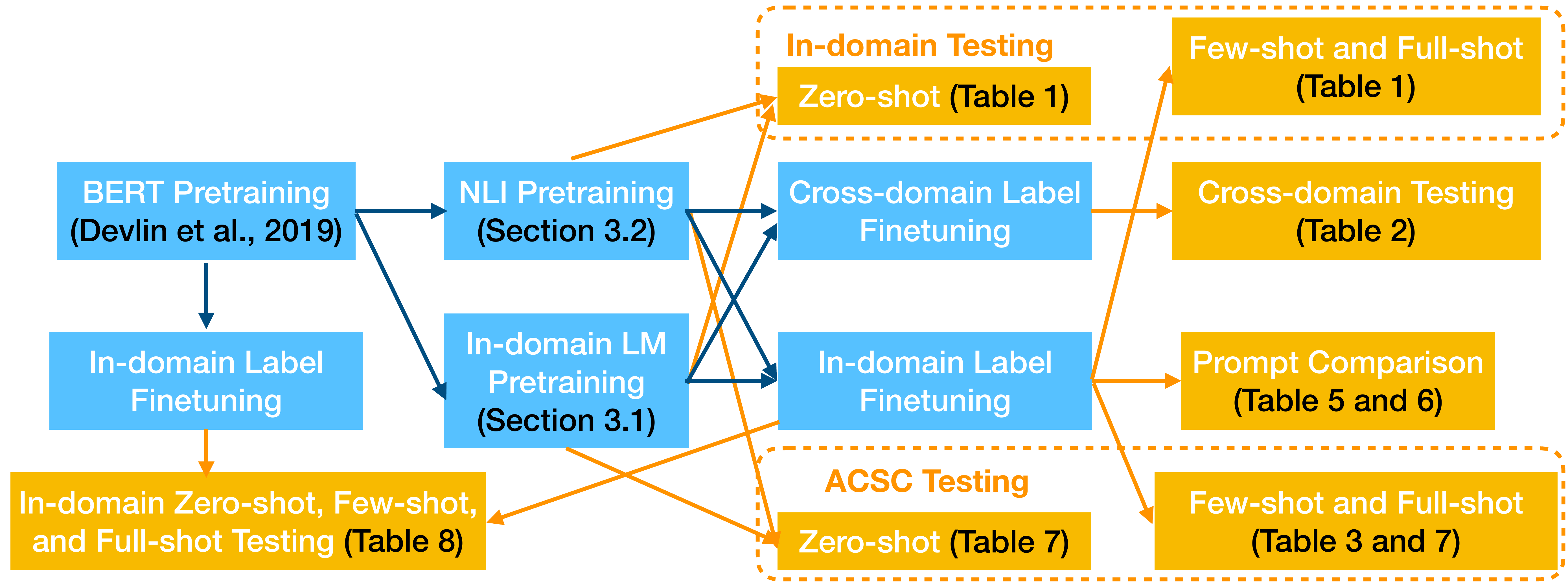}
\caption{The flow chart of our methods and experiments. The blue blocks represent the training steps and the orange blocks represent the testing steps.}
\label{fig:setting_summary}
\end{figure*}

\section{Training and Testing Settings}
\label{appendix:other_training}

Our training and testing settings are summarized in \autoref{fig:setting_summary}.

All the program codes used to produce results presented in this paper are available at \url{https://link.iamblogger.net/atscprompts}.

\paragraph{Publicly available BERT MLM, GPT-2 CLM, and BERT NLI weights} Following the common practice in recent NLP transfer learning literature, we use publicly available weights pretrained on large unlabeled corpora and task datasets for further training. For BERT MLM and GPT-2 CLM, we use the weights obtained from the \texttt{transformers} library \cite{wolf2020huggingfaces}. For BERT NLI, we use the weights released by \citet{morris2020textattack}, which were trained on the MNLI dataset \citep{N18-1101}.

\paragraph{BERT / GPT-2 main layer finetuning} Unlike more usual ways of using prompts where the main layers of language models are left frozen and do not get to see any updates, we simply leave them open for further finetuning. While this technically leads to more amount of computation, we anticipated that the cost would be negligible given that the amount of labeled data for the task is fairly small, especially for few-shot learning cases we are particularly interested in.

\paragraph{Train-test split and training hyperparameters} For the full-shot training, we use the entirety of the train split of the SemEval dataset following \citet{rietzler19}.

For ATSC training, we train with the SemEval datasets for 20 epochs. Following the recommendations regarding BERT finetuning made by \citet{mosbach2020stability}, we finetune all our prompt-based models and no-prompt baselines until the minimum training loss reaches near zero to achieve stable task performance, where the minimum losses are around 1e-07 and 1e-06.

For each ATSC model, we train them 5 times with different random seeds. Using different random seeds changes the data loading order, and the subset of training examples chosen for few-shot settings.

\paragraph{Hardwares and Softwares Used} For each ATSC model, we used one NVIDIA TITAN X GPU for training. The version 4.3.3 of \texttt{transformers} library \cite{wolf2020huggingfaces} is used with \texttt{pytorch} version 1.7.1. We also implemented all the loading scripts for our datasets to be compatible with the version 1.2.1 of the Huggingface \texttt{datasets} library\footnote{\url{https://huggingface.co/docs/datasets/master/}}. We have used the \texttt{spacy} library \citep{spacy2} for POS tagging, and \texttt{pytokenizations}\footnote{\url{https://github.com/tamuhey/tokenizations}} for tokenizer alignment.

\section{Comparing Performance of Different Prompts}
\label{appendix:prompts_comparison}

\begin{table}[h!]
    \begin{subtable}[h]{\columnwidth}
    \begin{adjustbox}{width=\columnwidth}
    \begin{tabular}{p{0.8\linewidth}cc}
        \toprule
        \multirow{2}{*}{\textbf{Prompt}}                               & \textbf{Accuracy}           & \textbf{Macro F1} \\
                                                        & \msa{\textbf{(Std. Error)}} & \msa{\textbf{(Std. Error)}}\\
        \midrule
        \multirow{2}{*}{ BERT NSP (No prompt) }                        & 48.24                       & 31.35\\
                                                        & \msa{(0.0283)}              & \msa{(0.0198)}\\
        \specialrule{.2em}{.3em}{.3em}
        \multirow{2}{*}{"I felt the \{aspect\} was [MASK]."}   & 69.06                       & 59.71\\
                                                        & \msa{(0.0060)}              & \msa{(0.0214)}\\
        \midrule
        \multirow{2}{*}{"The \{aspect\} made me feel [MASK]."}  & 68.15                       & 56.59\\
                                                        & \msa{(0.0069)}              & \msa{(0.0205)}\\
        \midrule
        \multirow{2}{*}{"The \{aspect\} is [MASK]." }           & 69.94                       & 59.51\\
                                                        & \msa{(0.0061)}              & \msa{(0.0179)}\\
        \bottomrule
    \end{tabular}
    \end{adjustbox}
    \caption{Laptops}
    \end{subtable}
    \begin{subtable}[h]{\columnwidth}
    \begin{adjustbox}{width=\columnwidth}
    \begin{tabular}{p{0.8\linewidth}cc}
        \toprule
        \multirow{2}{*}{\textbf{Prompt}}                               & \textbf{Accuracy}            & \textbf{Macro F1} \\
                                                        & \msa{\textbf{(Std. Error)}} & \msa{\textbf{(Std. Error)}}\\
        \midrule
        \multirow{2}{*}{BERT NSP (No prompt)}                          & 61.05          & 32.46\\
                                                        & \msa{(0.0238)} & \msa{(0.0374)} \\
        \specialrule{.2em}{.3em}{.3em}
        \multirow{2}{*}{"I felt the \{aspect\} was [MASK]." }   & 73.59          & 59.03\\
                                                        & \msa{(0.0247)} & \msa{(0.0168)} \\
        \midrule
        \multirow{2}{*}{"The \{aspect\} made me feel [MASK]."}  & 69.38          & 51.65 \\
                                                        & \msa{(0.0223)} & \msa{(0.0114)} \\
        \midrule
        \multirow{2}{*}{"The \{aspect\} is [MASK]."}            & 73.02          & 59.25 \\
                                                        & \msa{(0.0209)} & \msa{(0.0152)} \\
        \bottomrule
    \end{tabular}
    \end{adjustbox}
    \caption{Restaurants}
    \end{subtable}
    \caption{Comparing different prompts on 16-shot training of our prompt models and baselines.}
    \label{tab:similar_prompts}
\end{table}

While Table 1 shows the scores averaged over different prompts, the performances are very similar across different manual prompts we have chosen, as seen during the full-shot training for the laptops domain in \autoref{tab:similar_prompts}. We observed similar trend in the restaurant domain, and few-shot scenarios we have tested. This suggests that practically reasonable choices of prompts could still achieve good ATSC performance.

\section{Importance of Aspect Dependent Prompts}
\label{appendix:no_aspect}

\begin{table}[h!]
    \begin{subtable}[h]{\columnwidth}
    \begin{adjustbox}{width=\columnwidth}
    \begin{tabular}{p{0.8\linewidth}cc}
        \toprule
        {\textbf{Prompt}}                              & \textbf{Accuracy}            & \textbf{Macro F1} \\
        \midrule
        {"I felt the \{aspect\} was [good/bad]."}   & 77.77                        & 73.56\\
        {"I felt the \ul{things} were [good/bad]."} & 74.06                        & 70.69\\
                                                            & \mdo{-3.71}                  & \mdo{-2.88} \\
        \bottomrule
    \end{tabular}
    \end{adjustbox}
    \caption{Laptops}
    \end{subtable}
    \begin{subtable}[h]{\columnwidth}
    \begin{adjustbox}{width=\columnwidth}
    \begin{tabular}{p{0.8\linewidth}cc}
        \toprule
        { \textbf{Prompt}}                              & \textbf{Accuracy}            & \textbf{Macro F1} \\
        \midrule
        {"The {aspect} is [good/bad]."}         & 85.93                         & 79.18\\
        {"The \ul{things} are [good/bad]."}    & 74.80                         & 64.95\\
                                                       & \mdo{-11.13}                  & \mdo{-14.22} \\
        \bottomrule
    \end{tabular}
    \end{adjustbox}
    \caption{Restaurants}
    \end{subtable}
    \caption{Performance changes with aspect terms removed from the best performing prompt for our NLI model.}
    \label{tab:no_aspect}
\end{table}

We performed small sanity check experiments where we confirm that given aspect terms in the prompts are actually being utilized to produce correct predictions. We take the best performing prompts for our NLI model, and replace all aspect terms with \texttt{things} for all test examples in the full-shot setting, regardless of actual aspect terms. Then, the exact prompt wording would become global for all inputs. \autoref{tab:no_aspect} shows that this brings significant drops in performance for both test domains, showing that the prompt needs to be aspect dependent to produce accurate predictions.

\section{Detailed ACSC Results}
\label{appendix:more_on_acsc}

\begin{table*}[p]
    \centering
    \begin{adjustbox}{width=\textwidth}
    \begin{tabular}{ccc|cc|cc|cc|cc|cc}
        \toprule
        \multirow{3}{*}{\textbf{Model}} & \multicolumn{12}{c}{\textbf{Number of ATSC Examples}} \\
                                           & \multicolumn{2}{c}{Zero} & \multicolumn{2}{c}{16} & \multicolumn{2}{c}{64} & \multicolumn{2}{c}{256} & \multicolumn{2}{c}{1024} & \multicolumn{2}{c}{Full} \\
                                                       & Acc   & MF1   & Acc   & MF1   & Acc   & MF1   & Acc   & MF1   & Acc   & MF1   & Acc   & MF1 \\
        \midrule
        {\centering BERT NSP}                          & --    & --    & 64.73 & 33.22 & 81.05 & 56.77 & 84.60 & 70.88 & 85.28 & 74.52 & 82.45 & 70.91\\
        \specialrule{.2em}{.3em}{.3em}
        \multirow{1}{*}{\centering BERT LM}            & 76.40 & 50.11 & 76.67 & 56.77 & 82.75 & 64.26 & 84.70 & 68.31 & 86.28 & 74.80 & 84.31 & 74.14\\
                                                       & \msa{--} & \msa{--} & \mup{+11.94} & \mup{+23.55} & \mup{+1.70} & \mup{+7.49} & \mup{+0.1} & \mdo{-2.57} & \mup{+1.00} & \mup{+0.28} & \mup{+1.86} & \mup{+3.23}\\
        \midrule
        \multirow{1}{*}{\centering BERT NLI}           & 44.36 & 40.77 & 66.92 & 58.18 & 73.41 & 63.67 & 69.52 & 60.66 & 70.81 & 61.61 & 67.42 & 59.24\\
                                                       & \msa{--} & \msa{--} & \mup{+2.19} & \mup{+24.96} & \mdo{-7.64} & \mup{+6.90} & \mdo{-15.08} & \mdo{-10.22} & \mdo{-14.47} & \mdo{-12.91} & \mdo{-15.03} & \mdo{-11.66}\\
        \bottomrule
    \end{tabular}
    \end{adjustbox}
    \caption{Performance on ACSC test data with our ATSC prompt models and baselines. We use 5 random seeds for each of the prompts and baselines, and average their scores.}
    \label{tab:results_acsc}
\end{table*}

In \autoref{tab:results_acsc}, we can see that BERT NLI is generally performing worse than BERT LM and BERT NSP in most cases. Unlike BERT LM, its performance appears to be stagnating even with more ATSC training examples. This trend is very different from what we observed in the main ATSC results in \autoref{tab:results}, where BERT NLI maintained high performance advantages in few-shot cases. The most probable cause for BERT NLI's worse performance is that it cannot fully comprehend the entailment relationships expressed with domain-specific vocabularies - due to not having done in-domain text pretraining, it seems quite likely for BERT NLI that it cannot recognize the facts such as \texttt{fajita} being a sort of \emph{food}.

\section{Effectiveness of In-domain LM Pretraining}
\label{appendix:vanilla_bert}

\begin{table*}[p]
    \centering
    \begin{subtable}[h]{\textwidth}
    \begin{adjustbox}{width=\textwidth}
    \begin{tabular}{cccc|cc|cc|cc|cc|cc}
        \toprule
        \multirow{3}{*}{\textbf{Model}} & \multirow{3}{*}{\textbf{Pretraining Corpora}} & \multicolumn{12}{c}{\textbf{Number of Training Examples}} \\
                                        &                                               & \multicolumn{2}{c}{Zero} & \multicolumn{2}{c}{16} & \multicolumn{2}{c}{64} & \multicolumn{2}{c}{256} & \multicolumn{2}{c}{1024} & \multicolumn{2}{c}{Full} \\
                                        &               & Acc   & MF1   & Acc   & MF1   & Acc   & MF1   & Acc   & MF1   & Acc   & MF1   & Acc   & MF1 \\
        \midrule
        {\centering BERT NSP} & Original              & --    & --    & 45.74 & 30.25 & 50.88 & 36.81 & 69.69 & 63.21 & 76.14 & 70.64 & 77.96 & 73.24\\
                              & Original + Amazon     & --    & --    & 48.24 & 31.35 & 60.91 & 49.27 & 72.38 & 64.64 & 76.77 & 71.12 & 80.25 & 75.46\\
                              &                       &       &       & \mup{+2.5} & \mup{+1.10} & \mup{+10.03} & \mup{+12.46} & \mup{+2.69} & \mup{+1.43} & \mup{+0.63} & \mup{+0.48} & \mup{+2.29} & \mup{+2.22}\\
        \specialrule{.2em}{.3em}{.3em}
        {\centering BERT LM} & Original               & 59.20 & 38.42 & 65.25 & 55.82 & 70.54 & 63.30 & 73.33 & 66.86 & 76.67 & 71.73 & 77.61 & 73.06 \\
                             & Original + Amazon      & 63.58 & 46.17 & 69.05 & 58.60 & 72.80 & 65.54 & 76.59 & 70.65 & 79.30 & 74.80 & 81.10 & 76.83 \\
                             &                        & \mup{+4.38} & \mup{+7.75} & \mup{+3.80} & \mup{+2.78} & \mup{+2.26} & \mup{+2.24} & \mup{+3.26} & \mup{+3.79} & \mup{+2.63} & \mup{+3.07} & \mup{+3.49} & \mup{+3.77}\\
        \bottomrule
    \end{tabular}
    \end{adjustbox}
    \caption{Laptops}
    \end{subtable}
    
    \hspace{100em}
    
    \begin{subtable}[h]{\textwidth}
    \begin{adjustbox}{width=\textwidth}
    \begin{tabular}{cccc|cc|cc|cc|cc|cc}
        \toprule
        \multirow{3}{*}{\textbf{Model}} & \multirow{3}{*}{\textbf{Pretraining Corpora}} & \multicolumn{12}{c}{\textbf{Number of Training Examples}} \\
                                        &                                               & \multicolumn{2}{c}{Zero} & \multicolumn{2}{c}{16} & \multicolumn{2}{c}{64} & \multicolumn{2}{c}{256} & \multicolumn{2}{c}{1024} & \multicolumn{2}{c}{Full} \\
                                        &               & Acc   & MF1   & Acc   & MF1   & Acc   & MF1   & Acc   & MF1   & Acc   & MF1   & Acc   & MF1 \\
        \midrule
        {\centering BERT NSP} & Original              & --    & --    & 55.00 & 34.09 & 63.36 & 37.26 & 74.39 & 58.16 & 80.57 & 70.09 & 84.77 & 76.93\\
                              & Original + Yelp     & --    & --    & 61.05 & 32.46 & 74.73 & 53.00 & 79.34 & 65.51 & 83.61 & 74.15 & 87.09 & 79.98\\
                              &                       &       &       & \mup{+6.05} & \mdo{-1.63} & \mup{+11.37} & \mup{+15.74} & \mup{+4.95} & \mup{7.35} & \mup{+3.04} & \mup{+4.06} & \mup{+2.32} & \mup{+3.05}\\
        \specialrule{.2em}{.3em}{.3em}
        {\centering BERT LM} & Original               & 68.04 & 36.44 & 65.73 & 51.93 & 74.88 & 58.21 & 77.24 & 63.76 & 81.82 & 72.21 & 84.26 & 75.90 \\
                             & Original + Yelp     & 70.86 & 48.17 & 71.99 & 56.65 & 77.79 & 63.30 & 81.10 & 69.27 & 85.12 & 76.60 & 87.50 & 80.78 \\
                             &                        & \mup{+2.82} & \mup{+11.73} & \mup{+6.26} & \mup{+4.72} & \mup{+2.91} & \mup{+5.09} & \mup{+3.86} & \mup{+5.51} & \mup{+3.30} & \mup{+4.39} & \mup{+3.24} & \mup{+4.88}\\
        \bottomrule
    \end{tabular}
    \end{adjustbox}
    \caption{Restaurants}
    \end{subtable}
    \caption{Comparing our prompt model performance between the original pretrained weights (``Original") and the weights further trained with domain-specific review texts (``Amazon, Yelp").}
    \label{tab:vanilla_vs_further}
\end{table*}

In \autoref{tab:vanilla_vs_further}, we show both BERT NSP and BERT LM results with the original pretrained weights (``Original") the weights further trained with domain-specific review texts (``Amazon, Yelp"). We could see that BERT LM with the original weights performs better than BERT NSP with domain-specific further pretraining in few shot settings. As previously suggested in \citet{scao2021many}, it appears that just using prompts alone does bring the positive benefits of alleviating the labeled data scarcity.
We also note that the gap between the original weights and the further pretrained ones is relatively small when more number of labeled examples becomes available for training.

\begin{table*}[p]
    \centering
    \begin{adjustbox}{width=0.9\textwidth}
    \begin{tabular}{cp{0.5\linewidth}ccccc}
        \toprule
        \textbf{Type} &\textbf{Review} & \textbf{Truth} & \textbf{Baseline} & \textbf{NLI} & \textbf{MLM} & \textbf{GPT-2}\\
        \midrule
        R1 &
        the good place to hang out during the day after shopping or to grab a simple soup or classic french dish over a \underline{\textit{glass of wine}}.
            & Neu
            & \color{redbw}Pos  
            & \color{green}Neu 
            & \color{green}Neu 
            & \color{green}Neu \\
        \midrule
        R2 &
        My friend had a \underline{\textit{burger}} and I had these wonderful blueberry pancakes.  
            & Neu
            & \color{redbw}Pos  
            & \color{green}Neu 
            & \color{green}Neu 
            & \color{green}Neu \\
        \midrule
        R3 &
        The \underline{\textit{sushi}} is cut in blocks bigger than my cell phone.
            & Neg
            & \color{redbw}Neu 
            & \color{green}Neg  
            & \color{redbw}Neu  
            & \color{redbw}Neu \\
        \midrule
        R4 &
        The absolute worst service I've ever experienced and the food was below average (when they actually gave people the \underline{\textit{meals}} they ordered).
            & Neu
            & \color{green}Neu 
            & \color{redbw}Neg  
            & \color{redbw}Neg  
            & \color{redbw}Neg \\
        \midrule
        R5 &
        \underline{\textit{Food}} was decent, but not great.
            & Pos
            & \color{redbw}Neu 
            & \color{redbw}Neu  
            & \color{redbw}Neu  
            & \color{redbw}Neu \\
        \bottomrule
    \end{tabular}
   \end{adjustbox}
    \caption{Analysis of various scenarios where our models and baselines fail. Aspect words are underlined, predictions are highlighted in green (correct) or red (incorrect).}
    \label{tab:errors_4}
\end{table*}

\section{Further Error Analysis}
\label{appendix:error_analysis}

\begin{table}[H]
    \centering
    \begin{adjustbox}{width=\columnwidth}
    \begin{tabular}{ccccccccc}
        \toprule
        \textbf{Model}  && \multicolumn{3}{c}{\textbf{Restaurants}}   && \multicolumn{3}{c}{\textbf{Laptops}}   \\
                        &&   Pos   &   Neg   &   Neu   &&    Pos   &   Neg   &   Neu   \\
        \midrule
        BERT NSP        &&   75.50 &  14.47   &  7.40   &&   63.65 &   22.76 & 7.64 \\
        \midrule
        BERT LM         &&   86.34 &  57.65   &  25.95   &&  83.51 &   60.83 & 31.46\\
        GPT-2 LM        &&   87.95 &  65.68   &  26.54   &&  82.85 &   66.83 & 20.47\\
        BERT NLI        &&   80.86 &  66.30   &  51.55   &&  83.60 &   69.60 & 50.98\\
        \bottomrule
    \end{tabular}
    \end{adjustbox}
    \caption{F1 scores for each class achieved by the baseline and our models with 16 examples.}
    \label{tab:class_results}
\end{table}

In \autoref{tab:class_results}, we show F1 scores for each classes. Over the baseline BERT NSP, all our prompt models show large improvements. Particularly with BERT NLI, we see that F1 for the neutral class has greatly improved, doing better than both BERT NSP and BERT LM. It appears that BERT NLI is particularly better than BERT LM at detecting neutral and negative examples, quite possibly because the MNLI dataset contains many examples with neutral and negative labels.

In \autoref{tab:errors_4}, we present a few notable examples from test data that one or more of our prompt-based models had predicted correctly while the baseline did not. R1 shows an example, which our model correctly classifies as neutral while the no-prompt baseline wrongly predicts positive. R2 shows an example where our models are able to make the correct prediction despite having multiple aspects within one sentence. We found R3 quite interesting, where there are no explicit terms to express negative sentiment, intuitively making it difficult for the model to detect sentiment; Yet only the NLI model is able to make the correct prediction. For future studies, it would be an interesting direction to perform further statistical analysis on whether phenomena we see here generally hold for prompt-based models and strong linguistic patterns emerge among them.

We also note some examples that our models were not able to classify correctly. For R4, all the models got wrong except for the baseline. This example seems particularly challenging, as there is another sentence showing negative sentiment about a very similar aspect (``the food was below average") R5 shows an example where the baseline and all our models fail — we find the true label here somewhat questionable, as intuitively the reviewer indeed appears to be neutral about food and not particularly positive. 

\end{document}